\def\papertitle{Universal audio synthesizer control with normalizing flows}
\def\paperauthorA{Philippe Esling\textsuperscript{1}, Naotake Masuda\textsuperscript{1}, Adrien Bardet\textsuperscript{1}, Romeo Despres\textsuperscript{1}, Axel Chemla--Romeu-Santos\textsuperscript{1,2}}

\documentclass[twoside,a4paper]{article}
\usepackage{dafx_19}
\usepackage{amsmath,amssymb,amsfonts,amsthm}
\usepackage{euscript}
\usepackage[latin1]{inputenc}
\usepackage[T1]{fontenc}
\usepackage{ifpdf}
\usepackage[ruled,linesnumbered]{algorithm2e}
\usepackage{multirow}
\usepackage{multicol}
\usepackage{makecell}
\usepackage{hyperref}

\usepackage[english]{babel}
\usepackage{caption}
\usepackage{subfig} 
\usepackage{color}

\usepackage{times}

\newif\ifpdf
\ifx\pdfoutput\relax
\else
   \ifcase\pdfoutput
      \pdffalse
   \else
      \pdftrue
\fi

\ifpdf 
  \usepackage[pdftex]{graphicx}
  \usepackage[figure,table]{hypcap}
\else 
  \usepackage[dvips]{epsfig,graphicx}
  \usepackage[figure,table]{hypcap}
\fi

\newcommand{\bb}[1]{\mathbf{#1}}

\newcommand{\bx}{\bb{x}}

\newcommand{\bt}{\bb{t}}

\newcommand{\bv}{\bb{v}}

\newcommand{\bz}{\bb{z}}

\newcommand{\bpsi}{\boldsymbol{\psi}}

\newcommand{\bepsilon}{\boldsymbol{\epsilon}}

\newcommand{\btheta}{\boldsymbol{\theta}}

\newcommand{\E}{\mathbb{E}}
\newcommand{\KL}[2]{\mathcal{D}_{\textsc{KL}}\left[#1 \| #2\right]}

\title{\papertitle}

\affiliation{
\paperauthorA }
{ \begin{multicols}{2}  \textsuperscript{1} IRCAM - CNRS UMR 9912 \\ Sorbonne Universit\'{e}, Paris, France \\ { \tt \href{mailto:esling@ircam.fr}{esling@ircam.fr}}  \\ \columnbreak \textsuperscript{2} Laboratorio d'Informatica Musicale (LIM) \\ UNIMI, Milano, Italy \\ { \tt \href{mailto:Axel.Chemla@unimi.it}{axel.chemla@unimi.it}} \\  \end{multicols}  }

\begin{document}
\ifpdf 
  \DeclareGraphicsExtensions{.png,.jpg,.pdf}
\else  
  \DeclareGraphicsExtensions{.eps}
\fi

\maketitle

\begin{abstract}
The ubiquity of sound synthesizers has reshaped music production and even entirely defined new music genres. However, the increasing complexity and number of parameters in modern synthesizers make them harder to master. Hence, the development of methods allowing to easily create and explore with synthesizers is a crucial need.

Here, we introduce a novel formulation of audio synthesizer control. We formalize it as finding an organized latent audio space that represents the capabilities of a synthesizer, while constructing an invertible mapping to the space of its parameters. By using this formulation, we show that we can address simultaneously \textit{automatic parameter inference}, \textit{macro-control learning} and \textit{audio-based preset exploration} 
within a single model. To solve this new formulation, we rely on Variational Auto-Encoders (VAE) and Normalizing Flows (NF) to organize and map the respective \textit{auditory} and \textit{parameter} spaces. We introduce the \textit{disentangling flows}, which allow to perform the invertible mapping between separate latent spaces, while steering the organization of some latent dimensions to match target variation factors by splitting the objective as partial density evaluation. We evaluate our proposal against a large set of baseline models and show its superiority in both parameter inference and audio reconstruction. We also show that the model disentangles the major factors of audio variations as latent dimensions, that can be directly used as \textit{macro-parameters}. We also show that our model is able to learn semantic controls of a synthesizer by smoothly mapping to its parameters. Finally, we discuss the use of our model in creative applications and its real-time implementation in Ableton Live\footnote{All code, supplementary figures, results and plugins are available on a supporting webpage: \href{https://acids-ircam.github.io/flow_synthesizer/}{https://acids-ircam.github.io/flow\_synthesizer/}}.

\end{abstract}

\section{Introduction}

Synthesizers are parametric systems able to generate audio signals ranging from musical instruments to entirely unheard-of sound textures. Since their commercial beginnings more than 50 years ago, synthesizers have revolutionized music production, while becoming
increasingly accessible, even to
neophytes with no background in signal processing.

While there exists a variety of sound synthesis types \cite{puckette2007theory}, they all require an \textit{a priori} knowledge to make the most out of a synthesizer possibilities. Hence, the main appeal of these systems (namely their versatility provided by large sets of parameters) also entails their major drawback. Indeed, the sheer combinatorics of parameter settings makes exploring all possibilities to find an adequate sound a daunting and time-consuming task. Furthermore, there are highly non-linear relationships between the parameters and the resulting audio. Unfortunately, no synthesizer provides intuitive controls related to perceptual and semantic properties of the synthesis. Hence, a method allowing an intuitive and creative exploration of sound synthesizers has become a crucial need, especially for non-expert users. 
\begin{figure}
\begin{center}
\includegraphics[scale=0.48]{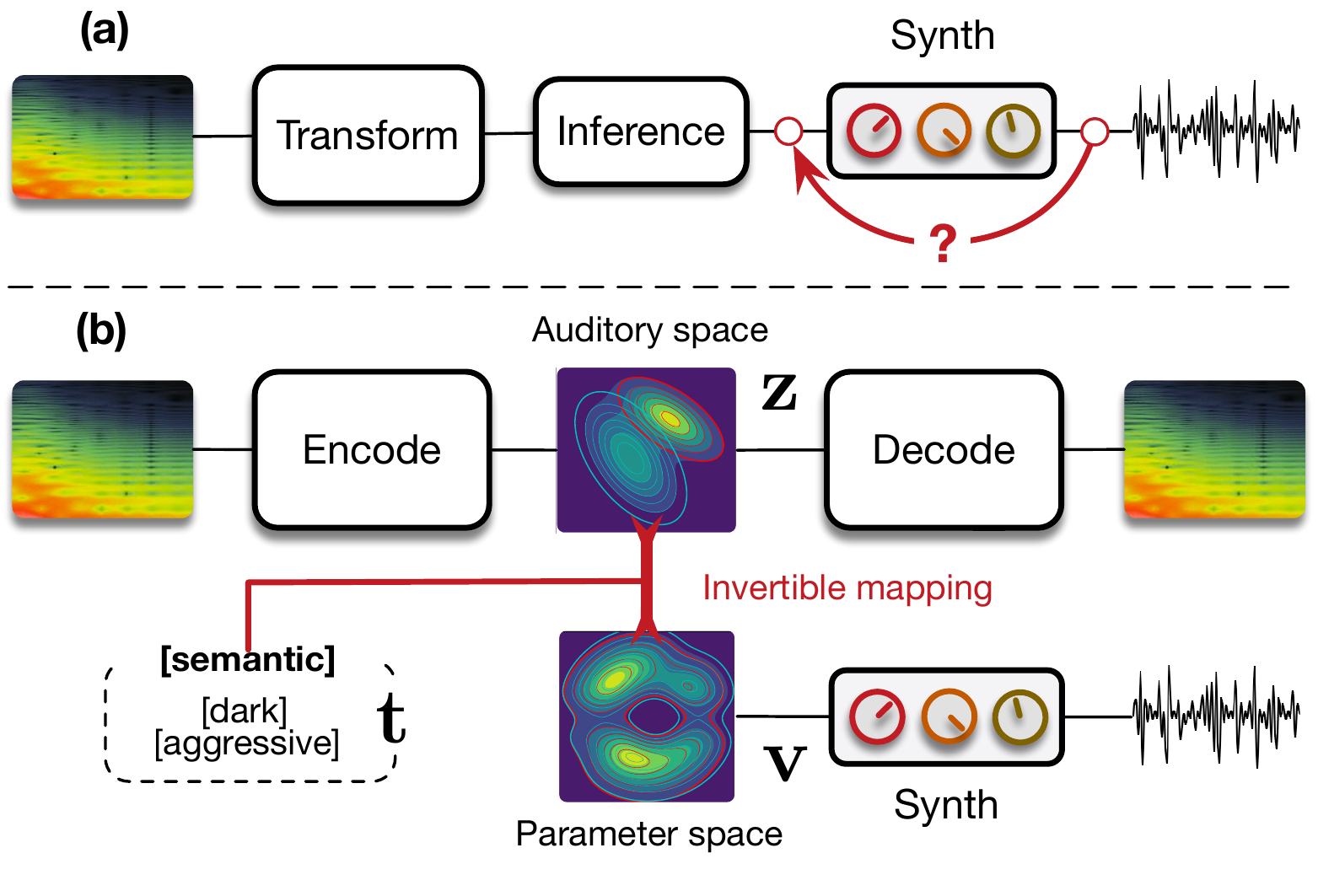}
\end{center}
\caption{\textit{Universal synthesizer control}. (a) Previous methods perform direct inference from audio, which is limited by non-differentiable synthesis and lacks high-level control. (b) Our novel formulation states allows to learn an organized latent space $\bz$ of the synthesizer's audio capabilities, while mapping it to the space $\bv$ of its synthesis parameters.}
\label{fig:workflow_compare}
\end{figure}


A potential direction taken by synth manufacturers, is to 
propose \textit{macro-controls} that allow to quickly tune a sound by controlling multiple parameters through a single knob. However, these need to be programmed manually, which still requires expert knowledge. Furthermore, no method has ever tried to tackle this \textit{macro-control learning} task, as this objective appears unclear and depends on a variety of unknown factors.
An alternative to manual parameter setting would be to infer the set of parameters that could best reproduce a given \textit{target sound}.
This task of \textit{parameter inference} has been studied in the past years using various techniques.
In Cartwright et al. \cite{cartwright2014synthassist_iterative_}, 
parameters are iteratively refined based on audio descriptors similarity and relevance feedback provided by the user. However, this approach appears to be rather inaccurate and slow. Garcia et al. \cite{garcia2002automatic} proposed to use genetic programming to directly grow modular synthesizers to solve this problem. Although the approach is appealing and appears accurate, the optimization of a single target can take from 10 to 200 hours, which makes it unusable. Recently, Yee-king et al. \cite{yee2018automatic} showed that a bi-directional LSTM with highway layers can produce accurate parameters appproximations. However, this approach does not allow for any user interaction. All of these approaches share the same flaws that (i) though it is unlikely that a synthesizer can generate exactly any audio target, none explicitly model these limitations, (ii) they do not account for the non-linear relationships that exist between parameters and the corresponding synthesized audio. Hence, no approach has succeeded in unveiling the true relationships between these \emph{auditory} and \emph{parameters} spaces. Here, we argue that it is mandatory to organize the parameters and audio capabilities of a given synthesizer in their respective spaces, while constructing an invertible mapping between these spaces in order to access a range of high-level interactions. This idea is depicted in Figure~\ref{fig:workflow_compare}


The recent rise of \textit{generative models} might provide an elegant solution to these questions. 
Indeed, amongst these models, the \textit{Variational Auto-Encoder} (VAE) \cite{kingma2013auto} aims to uncover the underlying structure of the data, by explicitly learning a \textit{latent space} \cite{kingma2013auto}. This space can be seen as a high-level representation, which aims to disentangle underlying variation factors and reveal interesting structural properties of the data \cite{kingma2013auto, higgins2016beta}. VAEs address the limitations of control and analysis through this latent space, while being able to learn on small sets of examples. Furthermore, the recently proposed \textit{Normalizing Flows} (NF) \cite{rezende2015variational} allow to model highly complex distributions in the latent space. Although the use of VAEs for audio applications has only been scarcely investigated, Esling et al. \cite{esling2018generative} recently proposed a perceptually-regularized VAE that learns a space of audio signals aligned with perceptual ratings via a regularization loss. The resulting space exhibits an organization that is well aligned with perception. Hence, this model appears as a valid candidate to learn an organized audio space. 

\begin{figure*}
\begin{center}
\includegraphics[scale=0.65]{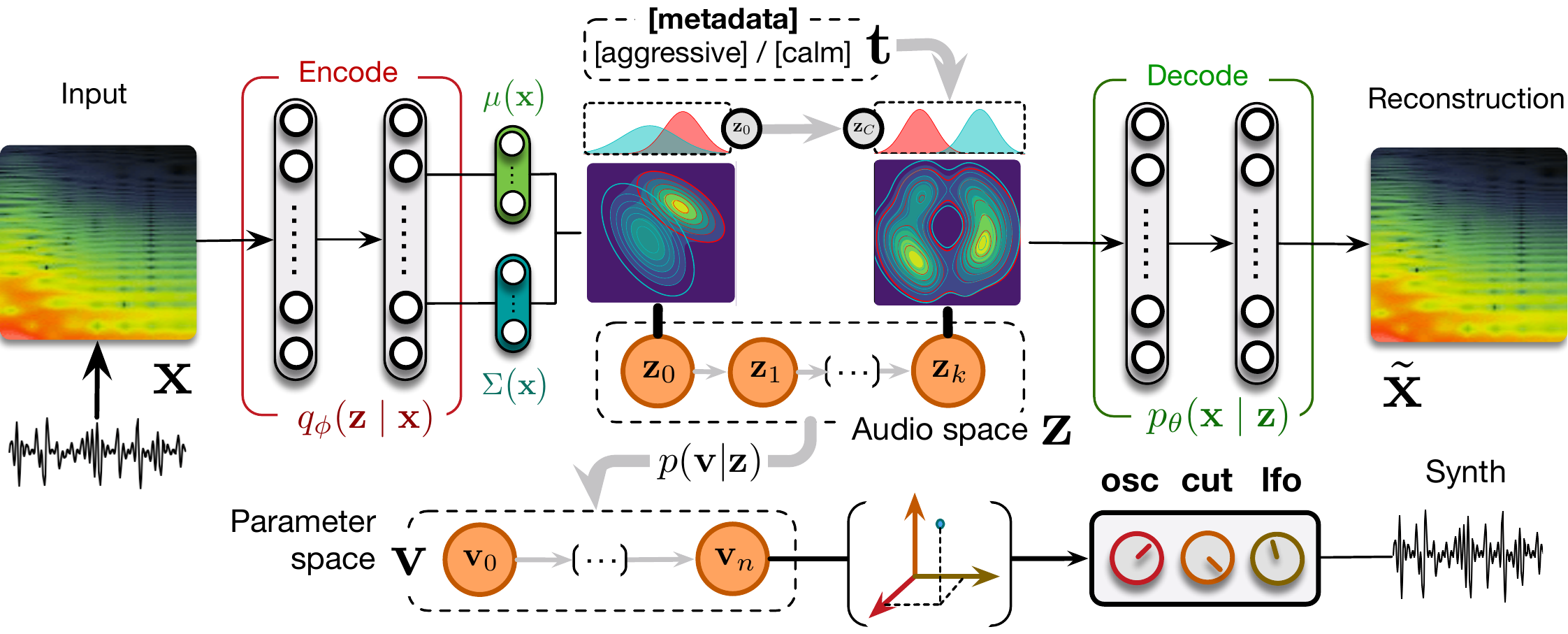}
\end{center}
\caption{\textit{Universal synthesizer control}. We learn an organized latent audio space $\bz$ of a synthesizer capabilities with a VAE parameterized with NF. This space maps to the parameter space $\bv$ through our proposed \textit{regression flow} and can be further organized with metadata targets $\bt$. This provides sampling and invertible mapping between different spaces. }
\label{fig:workflow_full}
\end{figure*}

In this paper, we introduce a radically novel formulation of audio synthesizer control by formalizing it as the general question of finding an invertible mapping between two learned latent spaces. In our case, we aim to map the audio space of a synthesizer's capabilities to the space of its parameters. We provide a generic probabilistic formalization and show that it allows to address simultaneously the tasks of \textit{parameter inference}, \textit{macro-control learning}, \textit{audio-based preset exploration} and \textit{semantic dimension discovery} within a single model. To elegantly solve this formulation, we introduce \textit{conditional regression flows}, which map a latent space to any given target space, while steering the organization of some dimensions to match target distributions. Our complete model is depicted in Figure~\ref{fig:workflow_full}.

Based on this formulation, \emph{parameter inference} simply consists of encoding the audio target to the latent audio space that is mapped to the parameter space. Interestingly, this bypasses the well-known blurriness issue in VAEs as we can generate directly with the synthesizer. We evaluate our proposal against a large set of baseline models and show its superiority in parameter inference and audio reconstruction. Furthermore, we show that our model is the first able to address the new task of automatic \textit{macro-control learning}. As the latent dimensions are continuous and map to the parameter space, they provide a natural way to learn the perceptually most significant macro-parameters. We show that these controls map to smooth, yet non-linear parameters evolution, while remaining perceptually continuous. Furthermore, as our mapping is invertible, we can map synthesis parameters back to the audio space. This allows intuitive \textit{audio-based preset exploration}, where exploring the neighborhood of a preset encoded in the audio space yields similarly sounding patches, yet with largely different parameters. Finally, we discuss creative applications of our model and real-time implementation in \textit{Ableton Live}.

\section{State-of-art}

\subsection{Generative models and variational auto-encoders}

\label{sec:generative_models_soa}

\textit{Generative models} aim to understand a given set $\mathbf{x}\in\mathbb{R}^{d_{x}}$ by modeling the underlying probability distribution of the data $p(\mathbf{x})$. To do so, we consider \emph{latent variables} defined in a lower-dimensional space $\mathbf{z}\in\mathbb{R}^{d_{z}}$ ($d_{z} \ll d_{x}$), a higher-level representation that could have led to generate a given example. The complete model is defined by the joint distribution $p(\mathbf{x}, \mathbf{z}) = p(\mathbf{x} \vert \mathbf{z})p(\mathbf{z})$. Unfortunately, real-world data follow complex distributions, which cannot be found analytically. The idea of \emph{variational inference} (VI) is to solve this problem through \emph{optimization} by assuming a simpler approximate distribution $q_{\phi}(\mathbf{z}\vert\mathbf{x})\in\mathcal{Q}$ from a family of approximate densities \cite{bishop2014pattern}. The goal of VI is to minimize the difference between this approximation and the real distribution, by minimizing the Kullback-Leibler (KL) divergence between these densities
$$
q_{\phi}^{*}(\mathbf{z}\vert \mathbf{x})=\text{argmin}_{q_{\phi}(\mathbf{z} \vert \mathbf{x})\in\mathcal{Q}} \mathcal{D}_{KL} \big[ q_{\phi}\left(\mathbf{z} \vert \mathbf{x}\right) \parallel p\left(\mathbf{z} \vert \mathbf{x}\right) \big]
$$
By developing this KL divergence and re-arranging terms (the detailed development can be found in \cite{kingma2013auto}), we obtain
\begin{multline}
    \log{p(\mathbf{x})} - D_{KL} \big[ q_{\phi}(\mathbf{z} \vert \mathbf{x}) \parallel p(\mathbf{z} \vert \mathbf{x}) \big] \\ 
= \mathbb{E}_{\mathbf{z}} \big[ \log{p(\mathbf{x} \vert \mathbf{z})}\big] - D_{KL} \big[ q_{\phi}(\mathbf{z} \vert \mathbf{x}) \parallel p(\mathbf{z}) \big]
\end{multline}
This formulation describes the quantity we want to model $\log p(\mathbf{x})$ minus the error we make by using an approximate $q$ instead of the true $p$. Therefore, we can optimize this alternative objective, called the \emph{evidence lower bound} (ELBO)
\begin{equation}
\mathcal{L}_{\theta, \phi} = \mathbb{E} \big[ \log{ p_\theta (\mathbf{x|z}) } \big] - \beta \cdot D_{KL} \big[ q_\phi(\mathbf{z|x}) \parallel p_\theta(\mathbf{z}) \big]
\end{equation}
The ELBO intuitively minimizes the reconstruction error through the likelihood of the data given a latent $\log{ p_\theta (\mathbf{x|z}) } $, while regularizing the distribution $q_{\phi}(\mathbf{z} \vert \mathbf{x})$ to follow a given prior distribution $p_{\theta}(\mathbf{z})$. We can see that this equation involves $q_{\phi}(\mathbf{z} \vert \mathbf{x})$ which \emph{encodes} the data $\mathbf{x}$ into the latent representation $\mathbf{z}$ and a \emph{decoder} $p_{\theta}(\mathbf{x} \vert \mathbf{z})$, which generates $\mathbf{x}$ given a $\mathbf{z}$. This structure defines the \emph{Variational Auto-Encoder} (VAE), where we can use parametric neural networks to model the \textit{encoding} ($q_{\phi}$) and \textit{decoding} ($p_{\theta}$) distributions. VAEs are powerful representation learning frameworks, while remaining simple and fast to learn without requiring large sets of examples \cite{sonderby2016train}.

However, the original formulation of the VAE entails several limitations. First, it has been shown that the KL divergence regularization can lead both to uninformative latent codes (also called \textit{posterior collapse}) and variance over-estimation \cite{chen2016variational}. One way to alleviate this problem is to rely on the \textit{Maximum Mean Discrepancy} (MMD) instead of the KL to regularize the latent space, leading to the WassersteinAE (WAE) model  \cite{tolstikhin2017wasserstein}. 
Second, one of the key aspect of VI lies in the choice of the family of approximations. The simplest choice is the \textit{mean-field} family where latent variables are mutually independent and parametrized by distinct variational parameters $ q(z) = \prod_{j=1}^{m}q_j(z_j)$. Although this provide an easy tool for analytical development, it might prove too simplistic when modeling complex data as this assumes pairwise independence among every latent axis. Normalizing flows alleviate this issue by adding a sequence of invertible transformations to the latent variable, providing a more expressive inference process.

\subsection{Normalizing flows}

In order to transform a probability distribution, we can rely on the \textit{change of variable} theorem. As we deal 
with probability distributions, we need to \textit{scale} the transformed density so that it still sums to one, which is measured by the determinant of the transform. Formally, let $\mathbf{z}\in\mathcal{R}^d$ be a random variable with distribution $q(\mathbf{z})$ and $f:\mathcal{R}^d\rightarrow\mathcal{R}^d$ an invertible smooth mapping. We can use $f$ to transform $\mathbf{z}\sim q(\mathbf{z})$, so that the resulting random variable $\mathbf{z}'=f(\mathbf{z})$ has the following probability distribution
\begin{equation}
q(\mathbf{z}')=q(\mathbf{z})\left|\text{ det}\frac{\partial f^{-1}}{\partial \mathbf{z}'}\right| = q(\mathbf{z})\left|\text{ det}\frac{\partial f}{\partial \mathbf{z}}\right|^{-1}
\end{equation}
where the last equality is obtained through the inverse function theorem. We can perform any number of transforms to obtain a final distribution $\mathbf{z}_k\sim q_k(\mathbf{z}_k)$ given by
\begin{align}
\begin{split}
q_k(\mathbf{z}_k) &= q_0(f_1^{-1} \circ ... \circ f_k^{-1}(\mathbf{z}_k))\prod_{i=1}^k\left|\text{det}\frac{\partial f^{-1}_i}{\partial\mathbf{z}_{i}}\right|\\
&= q_0(\mathbf{z_0})\prod_{i=1}^k\left|\text{det}\frac{\partial f_i}{\partial\mathbf{z}_{i-1}}\right|^{-1}
\end{split}
\end{align}
This series of transformations, called a \textit{normalizing flow} \cite{rezende2015variational}, can turn a simple distribution into a complicated multimodal density. 
For practical use of these flows, we need transforms whose Jacobian determinants are easy to compute. Interestingly, \textit{Auto-Regressive} (AR) transforms fit this requirement as they lead to a triangular Jacobian matrix. Hence, different AR flows were  proposed such as \textit{Inverse AR Flows} (IAF) \cite{kingma2016iaf} and \textit{Masked AR Flows} (MAF) \cite{papamakarios2017maf} 

\textit{Normalizing flows in VAEs}. Normalizing flows allow to address the simplicity of variational approximations by complexifying their posterior distribution \cite{rezende2015variational}. In the case of VAEs, we parameterize the approximate posterior distribution with a flow of length $K$, $q_{\phi}(\mathbf{z}\vert\mathbf{x}) = q_K(\mathbf{z}_K)$, and the new optimization loss can be simply written as an expectation over the initial distribution $q_0(\mathbf{z})$
\begin{align}
\begin{split}
\mathcal{L} &= \mathbb{E}_{q_{\phi}(\mathbf{z}\vert\mathbf{x})}\left[
\text{log }q_{\phi}(\bz\vert\mathbf{x}) - \text{log }p(\mathbf{x},\mathbf{z})\right]\\
&= \mathbb{E}_{q_0(\bz_0)}\left[\text{ln }q_{0}(\mathbf{z}_0)\right] - \mathbb{E}_{q_0(\bz_0)}\left[\text{log }p(\mathbf{x},\mathbf{z}_K)\right] \\ &- \mathbb{E}_{q_0(\bz_0)}\left[\sum_{i=1}^{k} \text{log} \left|\text{det}\frac{\partial f_i}{\partial\mathbf{z}_{i-1}}\right|\right] 
\end{split}
\end{align}
The resulting objective can be easily optimized since $q_0$ is still a Gaussian from which we can easily sample. However, the final samples $\mathbf{z}_k$ used by the decoder are drawn from a more complex distribution. 

\section{Our proposal}

\subsection{Formalizing synthesizer control}
\label{sec:formalization}

Considering a set of audio samples $\mathcal{D}=\left\{ \bx_i\right\}, i\in[1,n]$ where the $\bx_i\in\mathbb{R}^d$ follow an unknown distribution $p(\bx)$, we can define latent factors $\bz\in\mathbb{R}^{z}$ to model the joint distribution $p(\bx,\bz)=p(\bx\mid \bz)p(\bx)$ as detailed in Section~\ref{sec:generative_models_soa}. In our case, some $\bar{\bx} \in \mathcal{D}_s \subset \mathcal{D}$ inside this set have been generated by a given synthesizer. This synthesizer defines a generative function $f_{s}(\bv; p, i)=\bar\bx$ where $\bv\in\mathbb{R}^s$ is a set of parameters that produce $\bar{\bx}$ at a given pitch $p$ and intensity $i$. However, in the general case, we know that if $\bx_j\not\in\mathcal{D}_s$, then $\bx_j=f_s(\mathbf{v})+\mathbf{\epsilon}$ where $\mathbf{\epsilon}$ models the error made when trying to reproduce any audio $\bx_i$ with a given synthesizer. 
Finally, we consider that some audio examples are annotated with a set of \textit{categorical semantic tags} $\mathbf{t}_i=\{0,1\}^t$, which define high-level perceptual properties that separate \textit{unknown} latent factors $\bz$ and \textit{target} factors $\bt$. Hence, the complete generative story of a synthesizer can be defined as
\begin{equation}
     p(\bx,\bv,\bt,\bz) = p(\bx|\bv,\bt,\bz)p(\bv|\bt,\bz)p(\bt|\bz)p(\bz)
\end{equation}
This very general formulation entails our original idea that we should uncover the relationship between the latent audio $\bz$ and parameters $\bv$ spaces by modeling $p(\bv, \bz)$. The advantage of this formulation is that the reduced dimensionality $\mathbb{R}^z \ll \mathbb{R}^x$ of the latent $\bz$ simplifies the problem of parameters inference, by relying on a more adequate and smaller input space. Furthermore, this formulation also provides a natural way of learning \textit{macro-controls} by inferring $p(\bv | \bz)$ in the general case, where separate dimensions of $\bz$ are expected to produce smooth auditory transforms. 

\textit{Mapping latent spaces}. In order to map the latent $\bz$ and parameter $\bv$ spaces, we can first consider that $\bz$ and semantic tags $\bt$ are both unknown latent factors where $p(\bz')=p(\bz, \bt)$ so that we address the reduced problem
\begin{equation}
    \log p(\bx,\bv,\bz') = \log (p(\bx|\bv,\bz')p(\bz')) + \log p(\bv|\bz')
\end{equation}

This allows to separately model the variational approximation (Section~\ref{sec:generative_models_soa}), while solving the inference $p_\theta(\bv|\bz')$.

\subsection{Mapping latent spaces with regression flows}

In order to map the latent $\bz$ and parameter $\bv$ spaces, we first separate our formulation so that
\begin{equation}
    \log p_\theta(\bx,\bv,\bz) = \log (p_\theta(\bx|\bv,\bz)p_\theta(\bz)) + \log p_\theta(\bv|\bz)
\end{equation}

This allows to separately model the variational approximation detailed in Section~\ref{sec:generative_models_soa}, while solving separately the inference problem $p_\theta(\bv|\bz)$. To address this inference, we need to find the optimal parameters $\bpsi$ of a transform $f_{\bpsi}$ so that $\bv = f_{\bpsi} (\bz) + \bepsilon$, where $\bepsilon\sim \mathcal{N}(\mathbf{0},\mathbf{C}_v)$ models the inference error as a zero-mean additive Gaussian noise with covariance $\mathbf{C}_v$. Here, we assume that the covariance decomposes into $\mathbf{C}_v^{-1}=\sum_i exp(\lambda_i)\mathbf{Q}_i$, where $\mathbf{Q}_i$ are fixed basis functions and $\lambda$ are hyperparameters. Therefore, the full joint likelihood that we need to optimize is given by 
\begin{equation}
    \mathcal{L}_{f_{\bpsi}, \lambda}=\text{log}\left[p_\theta(\bv| f_{\bpsi}, \lambda, \bz)p_\theta(f_{\bpsi}|\bz)p_\theta(\lambda| \bz)\right]
\end{equation}

If we know the optimal transform $f_{\bpsi}$ and parameters $\lambda$, the likelihood of the data can be easily computed as
\begin{equation}
    p_\theta(\mathbf{v}\mid f_{\bpsi}, \lambda, \mathbf{z}) = \mathcal{N}(\mathbf{v}; f_{\bpsi}(\mathbf{z}), \mathbf{C}_v)
\end{equation}

However, the two posteriors $p_\theta(f_{\bpsi}|\bz)$ and $p_\theta(\lambda| \bz)$ remain intractable in the general case. In order to solve this issue, we rely again on variational inference by defining an approximation $q_\phi(f_{\bpsi}, \lambda | \bv, \bz)$ (see Section~\ref{sec:generative_models_soa}) and assume that it factorizes as $q(f_{\bpsi}, \lambda | \bv, \bz)=q(f_{\bpsi} | \bv, \bz)q(\lambda | \bv, \bz)$. Therefore, our final inference problem is 
\begin{align}
    \mathcal{L}_{f_{\bpsi}, \lambda} &= \text{log}\left[p_\theta(\bv| f_{\bpsi}, \lambda, \bz)\right] \nonumber\\
    &+ \KL{q_\phi(f_{\bpsi}|\bz,\bv)}{p_\theta(f_{\bpsi}|\bz)} \\
    &+ \KL{q_\phi(\lambda|\bz,\bv)}{p_\theta(\lambda|\bz)}
\end{align}
Hence, we can optimize our approximations through the KL divergence if we find a closed form. To solve for $\lambda$, we use a Gaussian distribution for both the prior $p_\theta(\lambda|\mathbf{z})=\mathcal{N}(\lambda, \mu_{\lambda}, C_{\lambda})$ and posterior $q_\phi(\lambda|\mathbf{z},\mathbf{v})=\mathcal{N}(\lambda, \mu_{q}, C_{q})$.
To solve this issue, we introduce the idea of \textit{regression flows}. This allows to obtain a simple analytical solution. However, the second part of the objective might be more tedious. Indeed, to perform an accurate inference, we need to rely on a complicated non-linear function, which cannot be assumed to be Gaussian. To address this issue, we introduce the idea of \textit{regression flows}. We consider that the transform $f_\theta(\bz)$ is a normalizing flow (see Section~\ref{sec:generative_models_soa}) and provides two different way of optimizing the approximation. 

\textit{Posterior parameterization}. 
First, we follow a reasoning akin to the original formulation of normalizing flows by parameterizing the posterior $q_\phi(f_{\bpsi}|\mathbf{z},\mathbf{v})$ with a flow $q_k(\bv_k)$. Hence, by developing the KL expression, we obtain
\begin{align}
    & \KL{q_\phi(f_{\bpsi}|\mathbf{z},\mathbf{v})}{p(f_{\bpsi}|\bz)} = \mathbb{E}_{q_0}\left[\text{log }q_{0}(\bv_0)\right] \nonumber \\ 
    &- \mathbb{E}_{q_0}\left[\text{log }p(\bv_k)\right] - \mathbb{E}_{q_0}\left[\sum_{i=1}^{k} \text{log} \left|\text{det}\frac{\partial f_i}{\partial\bv_{i-1}}\right|\right] 
\end{align}

Hence, we can now safely rely on Gaussian priors for $q_{0}(\bv_0)$ and $p(\bv_k)$. This formulation allows to consider $\bv$ as a transformed version of $\bz$, while being easily invertible as $\bz=f_{[k,1]}^{-1}(\bv)$. We denote this version as $Flow_{post}$.

\textit{Conditional amortization}.
Here, we consider that the parameters $\bpsi$ of the flow are random variables that are optimized by decomposing the posterior KL objective as
\begin{align}
    & \KL{q_\phi(f_{\bpsi}|\mathbf{z},\mathbf{v})}{p(f_{\bpsi}|\bz)} = \KL{q_\phi(\bpsi|\bz)}{p(\bpsi|\bz)}\nonumber \\ &+\mathbb{E}_{q_0(\bv_0)}\left[\sum_{i=1}^{k} \text{log} \left|\text{det}\frac{\partial f_i}{\partial\bv_{i-1}}\right|\right] 
\end{align}

As we rely on Gaussian priors for the parameters, this additional KL term can be computed easily. In this version, denoted $Flow_{cond}$, parameters of the flow are sampled from their distributions before computing the resulting transform.

\subsection{Disentangling flows for semantic dimensions}
\label{sec:disentangling}

We introduce \textit{semantic tags} in the training by expanding latent factors $\bz$ with categorical $\bt$. Hence, we define the generative process $p_\theta (\bx | \bt, \bz)$ where $p(\bt)=Cat(\bt|\bb{\pi})$ and $p(\pi)$ is the prior distribution of the tags. We define the inference model as $q_{\phi}(\bz, \bt | \bx)$ and assume that it factorizes as $q_{\phi}(\bz, \bt | \bx) = q_{\phi}(\bz | \bx)q_{\phi}(\bt | \bx)$. In order to handle the fact that tags are not always observed, we define a model similar to \cite{kingma2014semi}. When $\bt$ is unknown, it is considered as a latent variable over which we can perform posterior inference
\begin{align}
\mathcal{L}_u = & -\mathbb{E}\left[\log p_\theta(\bx|\bt,\bz) + \log p_\theta(\bt) + \log p_\theta(\bz)\right] \nonumber\\
& - \mathbb{E}\left[\log q_\phi(\bt,\bz|\bx)\right]
\end{align}

When tags $\bt$ are known, we take a rather unusual approach through the idea of \textit{disentangling flows}. As we seek to obtain a latent dimension with continuous semantic control, we define a tag pair as a set of negative $\bt_{-}$ and positive $\bt_{+}$ samples. We define two \textit{target} distributions $p(z_{\bt_{-}})\sim \mathcal{N}(-\mu_*, \sigma_{-})$ and $p(z_{\bt_{+}})\sim \mathcal{N}(+\mu_*, \sigma_{+})$ that model samples of a semantic pair as opposite sides of a latent dimension. Hence, we turn the treatment of tags into a \textit{density estimation} problem, where we aim to match tagged samples $\bt_{*}$ densities to targets minimizing $\KL{q_\phi(z_{\bt_{*}}|\bx)}{p(z_{\bt_{*}})}$. To solve this, we consider that $q_\phi(z_{\bt_{*}}|\bx)$ is parameterized by a normalizing flow $f_k$ applied to the latent $\bz$, leading to our final objective
\begin{align}
    &\mathcal{L}_{o} = \KL{q_\phi(z_{\bt_{*}})}{p(z_{\bt_{*}})} \nonumber \\
    &= \mathbb{E}\left[\log p(\bz)
    - \sum_{i=1}^{k} \text{log} \left|\text{det}\frac{\partial f_i}{\partial\mathbf{z}_{i-1}}\right|
    - \log p(z_{\bt_{*}})\right]
\end{align}

This formulation enforces a form of \textit{supervised} disentanglement, where latent $\bz$ are transformed to provide dimensions with explicit target properties. The final bound is defined as the sum of both objectives $\mathcal{L}=\mathcal{L}_o + \mathcal{L}_u$ and the complete model is obtained by integrating \textit{regression} and \textit{disentangling} flows together.

\section{Experiments}

\subsection{Dataset}

\emph{Synthesizer.}
We constructed a dataset of synthesizer sounds and parameters, by using an off-the-shelf commercial synthesizer \textit{Diva} developed by U-He\footnote{https://u-he.com/products/diva/}. It should be noted that our model can work for any synthesizer, as long as we obtain couples of (audio, parameters) as input. We selected Diva as (i) almost all its parameters can be MIDI-controlled, (ii) large banks of presets are available and (iii) presets include well-organized semantic tags pairs. The factory presets for Diva and additional presets from the internet were collected, leading to a total of roughly 11k files. We manually established the correspondence between synth and MIDI parameters as well as the parameters values range and distributions. We only kept continuous parameters and normalize all their values to $[0,1]$. All other parameters are set to their fixed \textit{default} value. Finally, we performed PCA and manual screening to select increasing sets of the most used 16, 32 and 64 parameters.  We use \textit{RenderMan}\footnote{https://github.com/fedden/RenderMan} to batch-generate all the audio files by playing the note for 3 sec. and recording for 4 sec. to capture the release of the note. The files are saved in 22050Hz and 16bit floating point format.

\emph{Audio processing.}
For each sample, we compute a 128 bins Mel-spectrogram with a FFT of size 2048 with a hop of 1024 and frequency range of $[30,11000]$. We only keep the magnitude of the spectrogram and perform a log-amplitude transform. The dataset is randomly split between a training (80\%), validation (10\%) and test (10\%) set before each training. We repeat the training $k$ times to perform $k$-fold cross-validation. Finally, we perform a corpus-wide zero-mean unit-variance normalization based on the train set.

%
%

\subsection{Models}

\emph{Baseline models.} In order to evaluate our proposal, we implemented several feed-forward deep models that take spectrograms $\bx_i$ as input and try to infer the corresponding parameters $\bv_i$. All these models are trained with a $MSE$ loss on the parameters vector. First, we implement a 5-layers $MLP$ with 2048 hidden units per layer, Exponential Linear Unit (ELU) activation, batch normalization and dropout with $p=.3$. This model is applied on a flattened version of the input and the final layer is a sigmoid activation. We implement a convolutional model composed of 5 layers with 128 channels of strided dilated 2-D convolutions with kernel size 7, stride 2 and an exponential dilation factor of $2^{l}$ (starting at $l=0$) with batch normalization and ELU activation. The convolutions are followed by a 3-layers MLP identical to the previous model. Finally, we implemented a \textit{Residual Network}, with parameters settings identical to $CNN$ and denote this model $ResCNN$\footnote{All remaining details on the models along with the complete source code for full reproducibility are available on the supporting webpage}.

\emph{Our models.} We implemented various *AE architectures with the $CNN$ setup for encoders and decoders. However, we halve their number of parameters (by dividing the number of units and channels) to perform a fair comparison by obtaining roughly the same capacity as the baselines. All models are trained with a $MSE$ reconstruction loss on the spectrograms. First, we implement a simple deterministic $AE$ without regularization. We implement the $VAE$ by adding a KL regularization to the latent space and the $WAE$ by replacing the KL by the MMD. Finally, we implement $VAE_{flow}$ by adding a normalizing flow of 16 successive IAF transforms to the $VAE$ posterior. All AEs map to latent spaces of dimensionality equal to the number of synthesis parameters. We perform \emph{warmup} \cite{sonderby2016train} by linearly increasing the latent regularization $\beta$ from 0 to 1 for 100 epochs. All AE models are trained with a 2-layers MLP to predict the parameters based on the latent space. Then, we use \textit{regression flows} ($Flow_{reg}$) by adding them to $VAE_{flow}$, with an IAF of length 16 without tags. Finally, we add the \textit{disentangling flows} ($Flow_{dis}$) by adding our objective defined in Sec.~\ref{sec:disentangling}

\emph{Optimization.} We train all models for 500 epochs with the ADAM optimizer, initial learning rate of 0.0002, Xavier initialization and a scheduler that halves the learning rate if the validation loss stalls for 20 epochs. With this setup, the most complex $VAE_{flow}$ with regression flows only needs ~5 hours to complete training on a NVIDIA Titan Xp GPU.

\begin{table*}
\begin{centering}
\scalebox{0.85}{
\begin{tabular}{r|c|cc||c|cc||cc}
\cline{2-9} 
\multicolumn{1}{r}{} & \multicolumn{3}{c|}{\textbf{Test set - 16 parameters}} & \multicolumn{3}{c||}{\textbf{Test set - 32 parameters}} & \multicolumn{2}{c}{\textbf{Out-of-domain} (32 p.)}\tabularnewline
\cline{2-9} 
\multicolumn{1}{r}{} & \multicolumn{1}{c|}{\textbf{Params}} & \multicolumn{2}{c||}{\textbf{Audio}} & \multicolumn{1}{c|}{\textbf{Params}} & \multicolumn{2}{c||}{\textbf{Audio}} & \multicolumn{2}{c}{\textbf{Audio}}\tabularnewline
\cline{2-9} 
\multicolumn{1}{r}{} & \multicolumn{1}{c|}{$MSE_{n}$} & \multicolumn{1}{c|}{$SC$} & \multicolumn{1}{c||}{$MSE$} & \multicolumn{1}{c|}{$MSE_{n}$} & \multicolumn{1}{c|}{$SC$} & \multicolumn{1}{c||}{$MSE$} & \multicolumn{1}{c|}{$SC$} & \multicolumn{1}{c}{$MSE$}\tabularnewline
\hline 
\hline 
$MLP$ & 0.236$\pm$.44 & 6.226$\pm$.13 & 9.548$\pm$3.1 & 0.218$\pm$.46 & 13.51$\pm$3.1 & 36.48$\pm$11.9 & 2.348$\pm$2.1 & 37.99$\pm$7.8\tabularnewline
$CNN$ & 0.171$\pm$.45 & 1.372$\pm$.29 & 6.329$\pm$1.9 & 0.159$\pm$.46 & 19.18$\pm$4.7 & 33.40$\pm$9.4 & 2.311$\pm$2.2 & 29.22$\pm$8.2\tabularnewline
$ResNet$ & 0.191$\pm$.43 & 1.004$\pm$.35 & 6.422$\pm$1.9 & 0.196$\pm$.49 & 10.37$\pm$1.8 & 31.13$\pm$9.8 & 2.322$\pm$1.6 & 31.07$\pm$9.5\tabularnewline
\hline 
$AE$ & 0.181$\pm$.40 & 0.893$\pm$.13 & 5.557$\pm$1.7 & 0.169$\pm$.40 & 5.566$\pm$1.2 & 17.71$\pm$6.9 & 1.225$\pm$2.2 & 27.37$\pm$7.2\tabularnewline
$VAE$ & 0.182$\pm$.32 & 0.810$\pm$.03 & 4.901$\pm$1.4 & 0.153$\pm$.34 & 5.519$\pm$1.4 & 16.85$\pm$6.1 & 1.237$\pm$1.3 & 27.06$\pm$7.1\tabularnewline
$WAE$ & \textbf{0.159$\pm$.37} & 0.787$\pm$.05 & 4.979$\pm$1.5 & \textbf{0.147$\pm$.33} & 3.967$\pm$.88 & 16.64$\pm$6.2 & 1.194$\pm$1.5 & 26.10$\pm$6.4\tabularnewline
$VAE_{flow}$ & 0.199$\pm$.32 & 0.838$\pm$.02 & 4.975$\pm$1.4 & 0.164$\pm$.34 & 1.418$\pm$.23 & 17.74$\pm$6.8 & 1.193$\pm$1.8 & 27.03$\pm$6.4\tabularnewline
\hline 
\hline 
$Flow_{reg}$ & 0.197$\pm$.31 & \textbf{0.752$\pm$.05} & \textbf{4.409$\pm$1.6} & 0.193$\pm$.32 & \textbf{0.911$\pm$1.4} & \textbf{16.61$\pm$7.4} & \textbf{1.101$\pm$1.2} & \textbf{26.07$\pm$7.7}\tabularnewline
\hline 
$Flow_{dis.}$ & 0.199$\pm$.31 & 0.831$\pm$.04 & 5.103$\pm$2.1 & 0.197$\pm$.42 & 1.481$\pm$1.8 & 17.12$\pm$7.9 & 1.209$\pm$1.4 & 26.77$\pm$7.3\tabularnewline
\hline 
\end{tabular}}
\par\end{centering}
\caption{Comparison between baselines, *AEs and our \textit{flows} on the test set with 16, 32 parameters and an out-of-domain set. We report across-folds mean and variance for parameters (MSE) and audio (SC and MSE) errors.} 
\label{tab:param_mapping}
\end{table*} 
\section{Results}

\begin{figure}
\begin{center}
\includegraphics[scale=0.41]{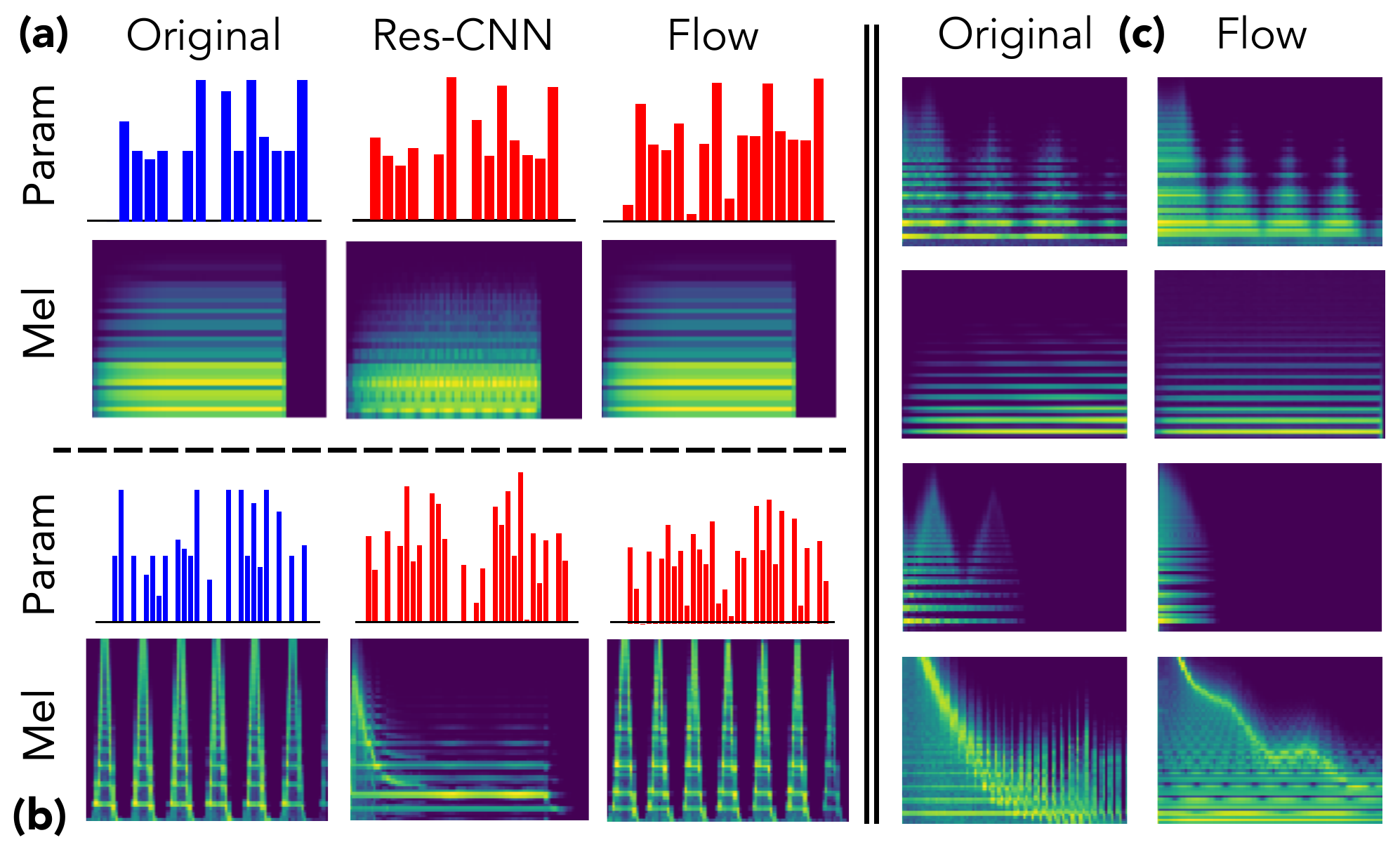}
\end{center}
\caption{\textit{Reconstruction analysis}. Comparing parameters inference and resulting audio on the test set with 16 (a) or 32 (b) parameters, and on the \textit{out-of-domain} (c) set.}
\label{fig:reconstruction}
\end{figure}

\subsection{Parameters inference}

First, we compare the accuracy of all models on \textit{parameters inference} by computing the magnitude-normalized \textit{Mean Square Error} ($MSE_n$) between predicted and original parameters values. We average these results across folds and report variance. We also evaluate the distance between the audio synthesized from inferred parameters and the original with the \textit{Spectral Convergence} (SC) distance (magnitude-normalized Frobenius norm) and \textit{MSE}. We provide evaluations for \textit{16} and \textit{32} parameters on the test set and an \textit{out-of-domain} dataset in Table~\ref{tab:param_mapping}.

In low parameters settings, baseline models seem to perform an accurate approximation of parameters, with the $CNN$ providing the best inference. Based on this criterion solely, our formulation would appear to provide only a marginal improvement, with $VAE$s even outperformed by baseline models and best results obtained by the $WAE$. However, analysis of the corresponding audio accuracy tells an entirely different story. Indeed, AEs approaches strongly outperform baseline models in audio accuracy, with the best results obtained by our proposed $Flow_{reg}$ (1-way ANOVA $F=2.81$, $p<.003$). These results show that, even though AE models do not provide an exact parameters approximation, they are able to account for the importance of these different parameters on the synthesized audio. This supports our original hypothesis that learning the latent space of synthesizer audio capabilities is a crucial component to understand its behavior. Finally, it appears that adding \text{disentangling flows} ($Flow_{dis}$) slightly impairs the audio accuracy. However, the model still outperform most approaches, while providing the huge benefit of explicit semantic macro-controls.

\textit{Increasing parameters complexity}. We evaluate the robustness of different models by increasing the number of parameters from 16 to 32. As we can see, the accuracy of baseline models is highly degraded, notably on audio reconstruction. Interestingly, the gap between parameter and audio accuracies is strongly increased. This seems logical as the relative importance of parameters in larger sets provoke stronger impacts on the resulting audio. Also, it should be noted that $VAE*$ models now outperform baselines even on parameters accuracy. Although our proposal also suffers from larger sets of parameters, it appears as the most resilient and manages this higher complexity. While the gap between AE variants is more pronounced, the \textit{flows} strongly outperform all methods ($F=8.13$, $p<.001$).

\textit{Out-of-domain generalization}. We evaluate \textit{out-of-domain generalization} with a set of audio samples produced by other synthesizers, orchestral instruments and voices. We rely on the same audio evaluation and provide results in Table~\ref{tab:param_mapping} (Right). Here, the overall distribution of scores remains consistent with previous observations. However, it seems that the average error is quite high, indicating a potentially distant reconstruction of some examples. 
Upon closer listening, it seems that the models fail to reproduce natural sounds (voices, instruments) but perform well with sounds from other synthesizers. In both cases, our proposal accurately reproduces the temporal shape of target sounds, even if the timbre is somewhat distant.


\subsection{Reconstructions and latent space}

We provide an analysis of the relations between inferred parameters and corresponding audio by selecting samples from different sets and displaying the results in Figure~\ref{fig:reconstruction}.

As we can see, although the $CNN$ provides a close inference of the parameters, the synthesized approximation misses important structural aspects, even in simpler instances of 16 parameters. This observation is amplified for 32 parameters, which confirms that direct inference models are unable to assess the \textit{relative impact} of parameters on the audio. Indeed, the error in all parameters is considered equivalently, even though the same error magnitude on two different parameters can lead to dramatic differences in the audio. Oppositely, while the parameters inferred by the VAE are quite far from the original, the corresponding audio is largely closer. This indicates that the latent space provides knowledge on \textit{audio-based neighborhoods}, allowing to understand the impact of different parameters in a given region of the latent audio space. Regarding \textit{out-of-domain} reconstruction, our proposal appears to provide an accurate rendition of the global temporal shape of the target audio, but seems to miss parts of the target timbre.

\subsection{Reconstructions and latent space}

We provide an in-depth analysis of the relations between inferred parameters and corresponding synthesized audio to support our previous claims. First, we selected two samples from the test set and compare the inferred parameters and synthesized audio in Figure~\ref{fig:reconstruction}.

\begin{figure}
\begin{center}
\includegraphics[scale=0.45]{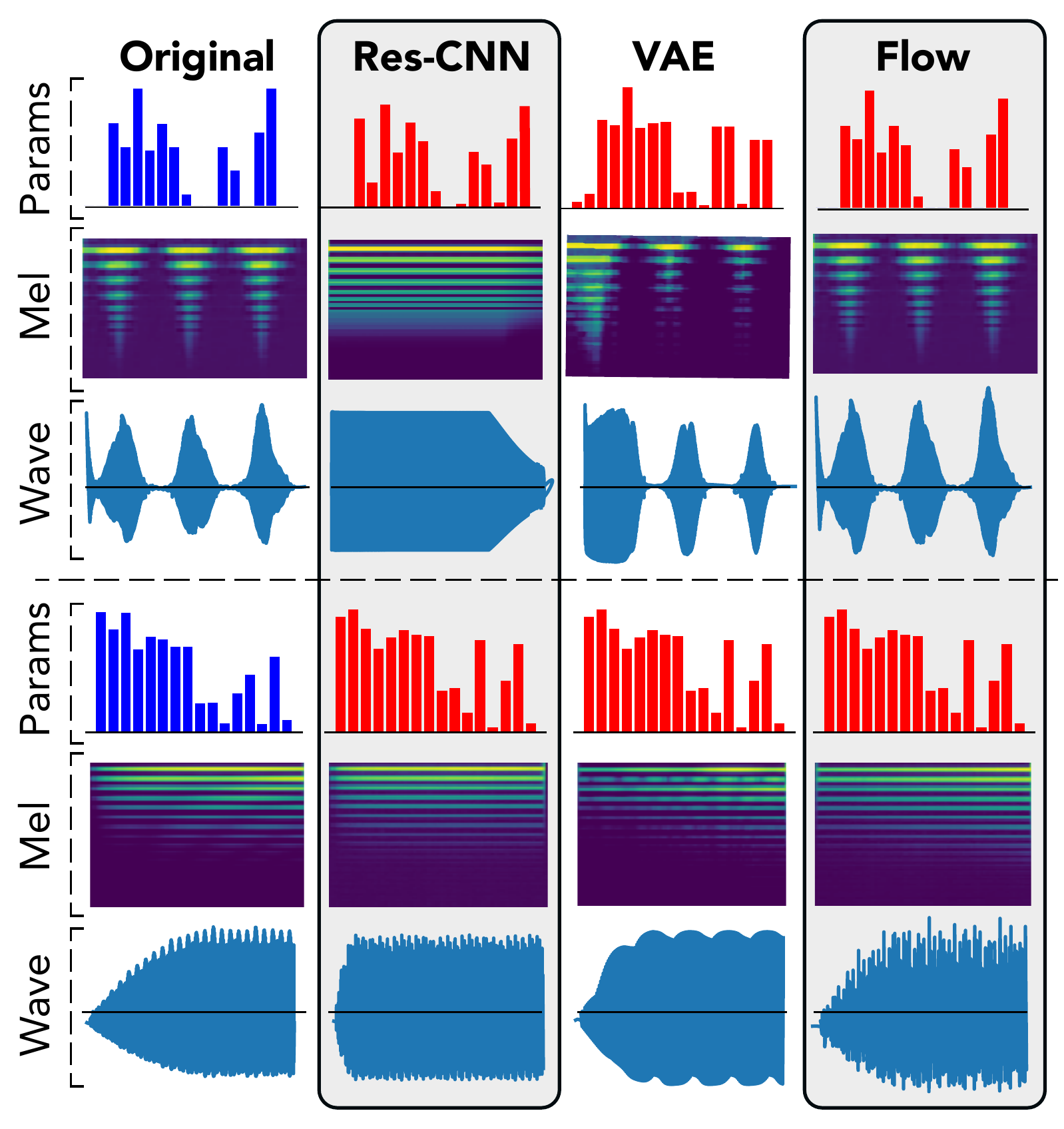}
\end{center}
\caption{\textit{Reconstruction analysis}. Comparing parameters inference and corresponding synthesized audio on the test dataset between the best performing models.}
\label{fig:reconstruction}
\end{figure}

As we can see, although the $CNN$ provides a close inference of the parameters, the synthesized approximation completely misses important structural aspects, even in simpler instances as the slow ascending attack in the second example. This confirms that direct inference models are unable to assess the \textit{relative impact} of parameters on the audio. Indeed, the errors in all parameters are considered equivalently, even though the same error magnitude on two different parameters can lead to dramatic differences in the synthesized audio. Oppositely, even though the parameters inferred by the VAE are quite far from the original preset, the corresponding audio is largely closer. This indicates that the latent space provides knowledge on the \textit{audio-based neighborhoods} of the synthesizer. Therefore, this allows to understand the impact of different parameters in a given region of the latent audio space.

\begin{figure}
\begin{center}
\includegraphics[scale=0.32]{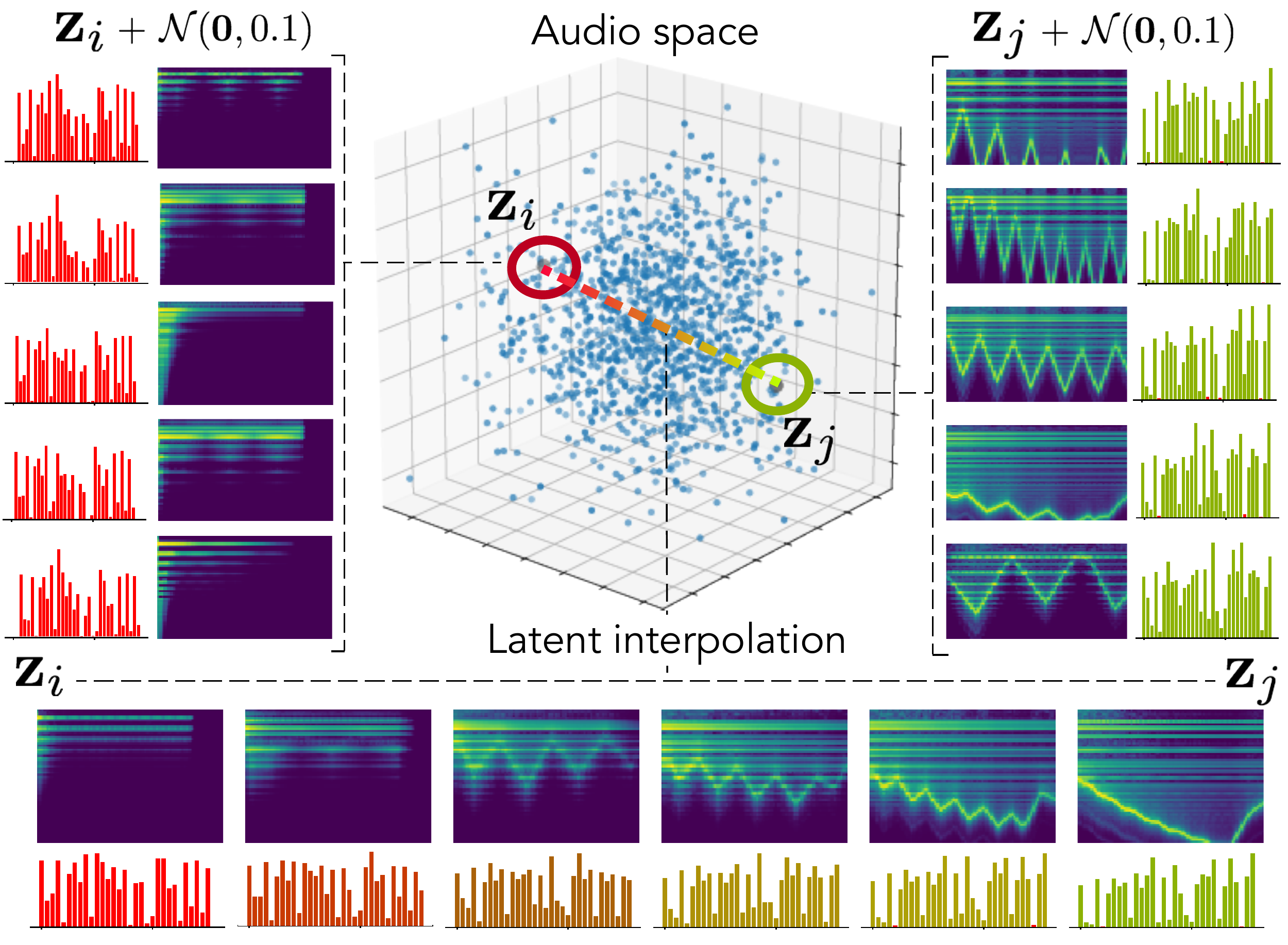}
\end{center}
\caption{\textit{Latent neighborhoods}. We select two examples from the test set that map to distant locations in the latent space $\bz$ and perform random sampling in their local neighborhood to observe the parameters and audio. We also display the latent interpolation between those points.}
\label{fig:latent_traversal}
\end{figure}

To evaluate this hypothesis, we encode two distant presets in the latent audio space and perform random sampling around these points to evaluate how local neighborhoods are organized. We also analyze the latent interpolation between those examples. The results are displayed in Figure~\ref{fig:latent_traversal}.

As we can see, our hypothesis seems to be confirmed by the fact that neighborhoods are highly similar in terms of audio but have a larger variance in terms of parameters. Interestingly, this leads to complex but smooth non-linear dynamics in the parameters control.

\subsection{Macro-parameters learning}

Our formulation is the first to provide a continuous mapping between the audio $\bz$ and parameter $\bv$ spaces of a synthesizer. As latent VAE dimensions has been shown to disentangle major data variations, we hypothesized that we could directly use $\bz$ as \textit{macro-parameters} defining the most interesting variations in a given synthesizer. Hence, we introduce the new task of \textit{macro-parameters learning} by mapping latent audio dimensions to parameters through $p(\bv|\bz)$, which provides simplified control of the major audio variations for a given synthesizer. This is depicted in Figure~\ref{fig:macro_params}

\begin{figure*}
\begin{center}
\includegraphics[scale=0.34]{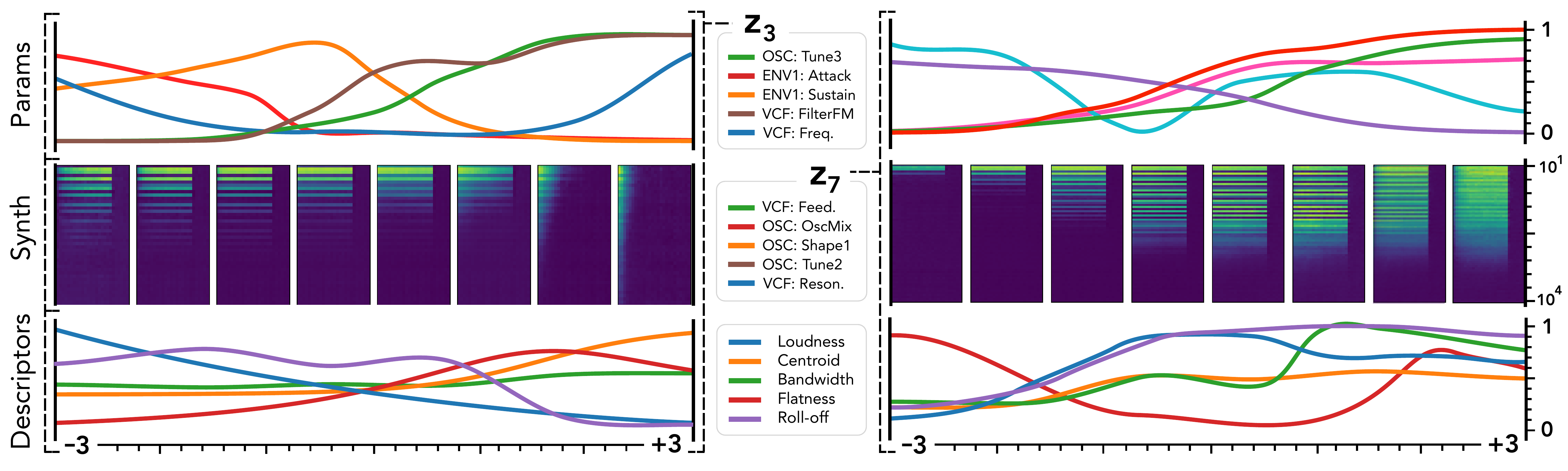}
\end{center}
\caption{\textit{Macro-parameters learning}. We show two of the learned latent dimensions $\bz$ and compute the mapping $p(\bv|\bz)$ when traversing these dimensions, while keeping all other fixed at $\bb{0}$ to see how $\bz$ define smooth macro-parameters. We plot the evolution of the 5 parameters with highest variance (top), the corresponding synthesis (middle) and audio descriptors (bottom). (Left) $\bz_3$ seems to relate to a \textit{percussivity} parameter. (Right) $\bz_7$ defines an \textit{harmonic densification} parameter.}
\label{fig:macro_params}
\end{figure*}

We show the two most informative latent dimensions $\bz$ based on their variance. We study the traversal of these dimensions by keeping all other fixed at $\bb{0}$ to assess how $\bz$ defines smooth macro-parameters through the mapping $p(\bv|\bz)$. We report the evolution of the 5 parameters with highest variance (top), the corresponding synthesis (middle) and audio descriptors (bottom). 

First, we can see that latent dimension corresponds to very smooth evolutions in terms of synthesized audio and descriptors. This is coherent with previous studies on the disentangling abilities of VAEs \cite{higgins2016beta}. However, a very interesting property appear when we map to the parameter space. Although the parameters evolution is still smooth, it exhibits more non-linear relationships between different parameters. This correlates with the intuition that there are lots of complex interplays in parameters of a synthesizer.
Our formulation allows to alleviate this complexity by automatically providing \textit{macro-parameters} that are the most relevant to the audio variations of a given synthesizer. Here, we can see that the $\bz_3$ latent dimension (left) seems to provide a \textit{percussivity} parameter, where low values produce a very slow attack, while moving along this dimension, the attack becomes sharper and the amount of noise increases. Similarily, $\bz_7$ seems to define an \textit{harmonic densification} parameter, starting from a single peak frequency and increasingly adding harmonics and noise. 

\subsection{Semantic parameter discovery}

Our proposed \textit{disentangling flows} can steer the organization of selected latent dimensions so that they provide a separation of given tags. As this audio space is mapped to parameters through $p(\bv|\bz)$, this turns the selected dimensions into \textit{macro-parameters} with a clearly defined semantic meaning. To evaluate this, we analyze the behavior of corresponding latent dimensions, as depicted in Figure~\ref{fig:macro_params}.

\begin{figure*}
\begin{center}
\includegraphics[scale=0.32]{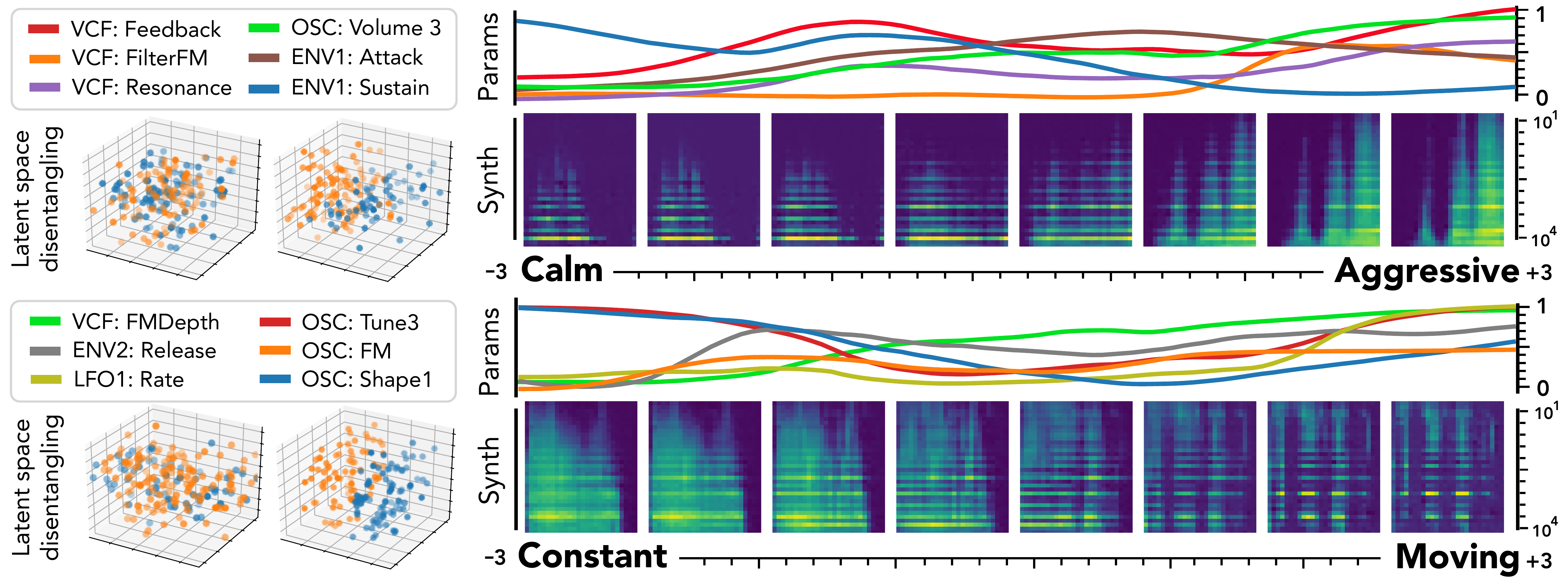}
\end{center}
\caption{\textit{Semantic macro-parameters}. Two latent dimensions $\bz$ learned through \textit{disentangling flows} for different pairs. We show the effect on the latent space (left) and parameters mapping $p(\bv|\bz)$ when traversing these dimensions, that define smooth macro-parameters. We plot the evolution of 6 parameters with highest variance and the resulting synthesized audio.}
\label{fig:macro_params}
\end{figure*}

First, we can see the effect of disentangling flows on the latent space (left), which provide a separation of semantic pairs. We study the traversal of semantic dimensions while keeping all other fixed at $\bb{0}$ and infer parameters through $p(\bv|\bz)$. We display the 6 parameters with highest variance and the resulting synthesized audio. As we can see, the semantic latent dimensions provide a very smooth evolution in terms of both parameters and synthesized audio. Interestingly, while the parameters evolution is smooth, it exhibits non-linear relationships between different parameters. This correlates with the intuition that there are complex interplays in parameters of a synthesizer. Regarding the effect of different semantic dimensions, it appears that the [\textit{'Constant'}, \textit{'Moving'}] pair provides a very intuitive result. Indeed, the synthesized sounds are mostly stationary in extreme negative values, but gradually incorporate clearly marked temporal modulations. Hence, our proposal appears successful to uncover \textit{semantic macro-parameters} for a given synthesizer. However, the corresponding parameters are quite harder to interpret. The [\textit{'Calm'}, \textit{'Aggressive'}] dimension also provides an intuitive control starting from a sparse sound and increasingly adding modulation, resonance and noise. However, we note that the notion of '\textit{Aggressive}' is highly subjective and requires finer analyses to be conclusive.

\subsection{Creative applications}

Our proposal allows to perform a direct exploration of presets based on audio similarity. Indeed, as the flow is \textit{invertible}, we can map parameters to the audio space for exploration, and then back to parameters to obtain a new preset. Furthermore, this can be combined with \textit{vocal sketch control} where the user inputs vocal imitations of the sound that he is looking for. This allows to quickly produce an approximation of the intended sound and then exploring the audio neighborhood of the sketch for intuitive refinement. We embedded our model inside a \textit{MaxMSP} external called \texttt{flow\_synth\~} by using the \textit{LibTorch} API and further integrate it into \textit{Ableton Live} by using the \textit{Max4Live} interface.

\section{Conclusion}

In this paper, we introduced several novel ideas including reformulating the problem of synthesizer control as matching the two latent space defined as the \textit{user perception space} and the \textit{synthesizer parameter space}.
We showed that our approach outperforms all previous proposals on the seminal problem of \textit{parameters inference}. Our formulation also naturally introduces the original tasks of
\emph{macro-control learning}, \emph{audio-based preset exploration} and \textit{semantic parameters discovery}. This proposal is the first to be able to simultaneously address most synthesizer control issues at once. 

Altogether, we hope that this work will provide new means of exploring audio synthesis, sparkling the development of new leaps in musical creativity.

\section{Acknowledgements}
This work was supported by MAKIMOno project (ANR:17-CE38-0015-01 and NSERC:STPG 507004-17) and the ACTOR Partnership (SSHRC:895-2018-1023).

\bibliographystyle{IEEEbib}
\bibliography{main} 



\end{document}